\documentclass[11pt]{article}

\usepackage[final]{acl}

\usepackage{times}
\usepackage{latexsym}
\usepackage[T1]{fontenc}
\usepackage[utf8]{inputenc}
\usepackage{microtype}
\usepackage{inconsolata}

\usepackage{booktabs}
\usepackage{graphicx}
\usepackage{subcaption}
\usepackage{xcolor}
\usepackage{colortbl}
\usepackage{multirow}
\usepackage{amsmath}
\usepackage{amsthm}

\usepackage{url}
\usepackage{hyperref}
\usepackage{cleveref}
\usepackage{xspace}
\usepackage{algorithm}
\usepackage[noend]{algpseudocode}
\usepackage{listings}
\usepackage[most]{tcolorbox}
\usepackage{tikz}
\usetikzlibrary{positioning, arrows.meta}

\lstdefinelanguage{json}{
  basicstyle=\ttfamily\footnotesize,
  showstringspaces=false,
  breaklines=true,
  numbers=none,
  morecomment=[l]{//},
  commentstyle=\color{black!50},
  moredelim=*[s][\color{red!80!black}]{<}{>}
}

\crefname{lstlisting}{Listing}{Listings}
\Crefname{lstlisting}{Listing}{Listings}

\title{What Else Needs Fixing? Exploring Cost-Effective Test-Time Compute for Revision Propagation in Artifacts Generated Through Conversation}
\author{Daisuke Kikuta \\
  NTT, Inc. \\
  \texttt{daisuke.kikuta@ntt.com}
}

\begin{document}
\maketitle

\begin{abstract}
Large Language Models (LLMs) often help users generate artifacts through iterative cycles of generation and revision in conversation.
A challenge here is that, when users specify only a local change during revision, LLMs must instead identify the relevant dependencies and propagate the revision to all affected parts of the artifact.
This paper studies this ability of LLMs on conversationally generated artifacts, where the artifact context and its dependencies may be embedded in the conversation history.
Toward practical use, we also explore cost-effective test-time compute for this new setting.
Specifically, we introduce a new benchmark for this setting, and evaluate nine revision methods, including sequential reflection and parallel sampling variants, using gpt-oss-20b/120b, gpt-5.4-mini, and qwen3.5-9b/27b/122b on the benchmark. 
The results show that baselines achieve accuracies of 68.3--93\%, and the most cost-effective method is selecting from three parallel samples using either LLM-based or medoid selection, which improves accuracy by 2.2--9.7\%.
Our code and dataset are available at \url{https://github.com/ntt-dkiku/llm-revision-propagation}.
\end{abstract}

\section{Introduction}
\begin{figure}[t]
\centering
\includegraphics[width=\linewidth]{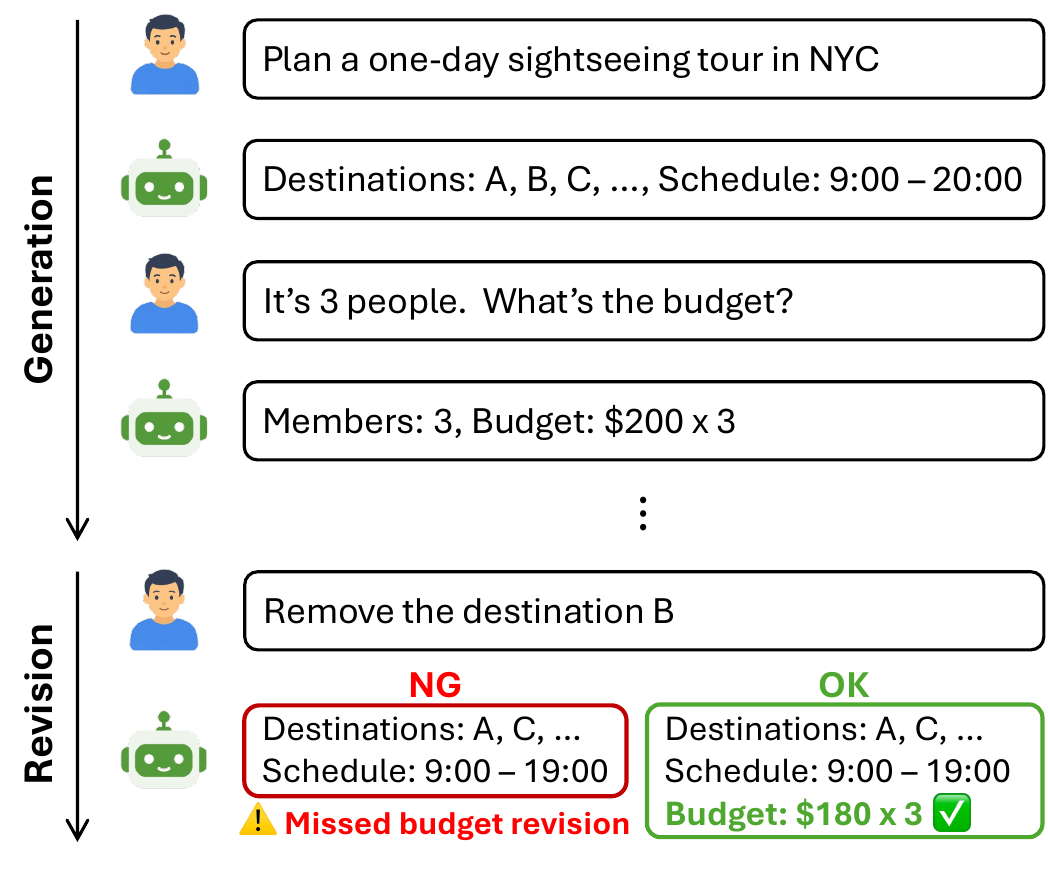}
\caption{Illustration of revision propagation in conversationally generated artifacts. Given a local revision request, LLMs must identify implicit dependencies and update not only the explicitly mentioned element but also other dependent elements to preserve consistency.}
\label{fig:teaser}
\end{figure}

\begin{figure*}[t]
\centering
\includegraphics[width=\linewidth]{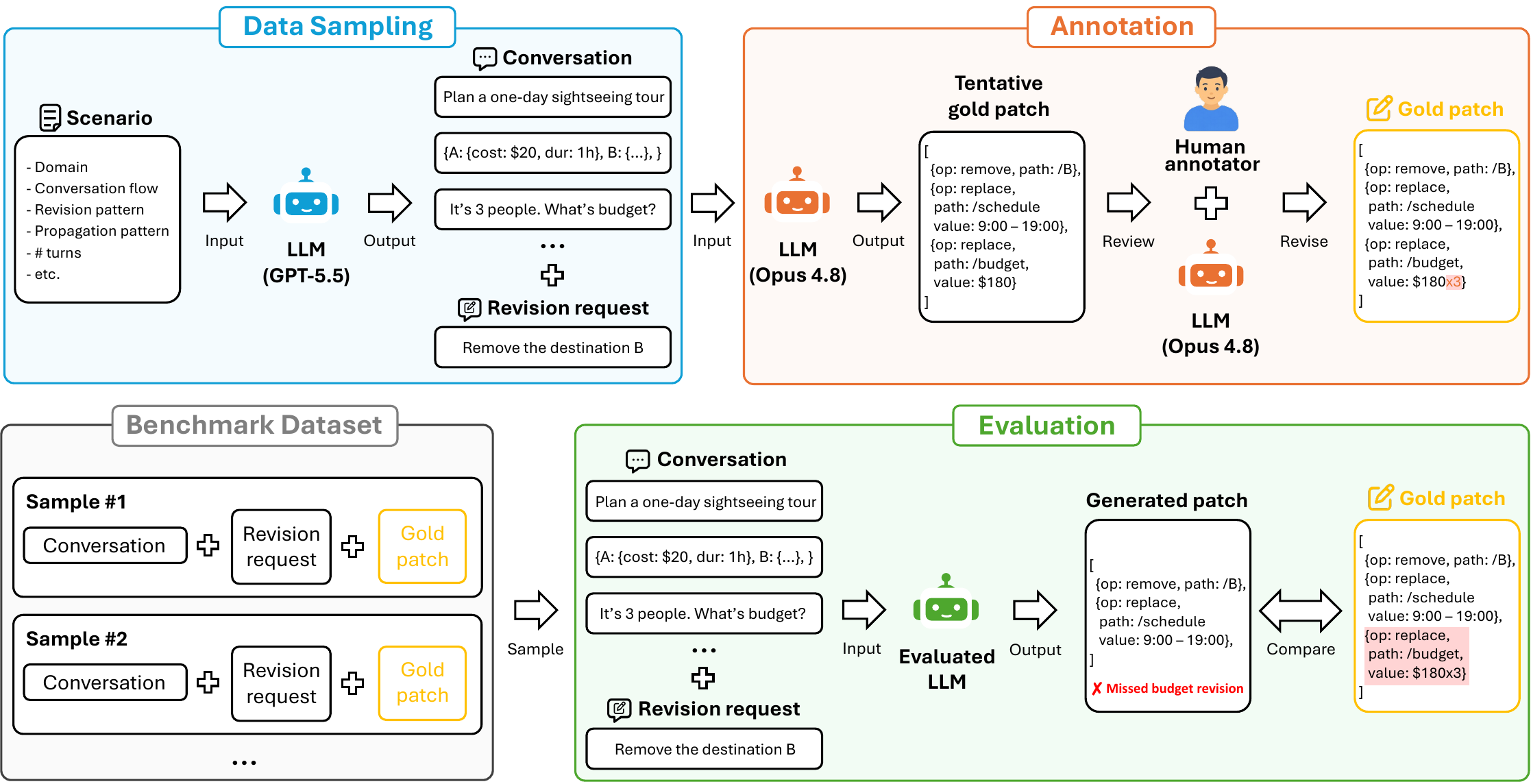}
\caption{Overview of RevPropBench. The upper half shows the benchmark construction process, while the lower half shows the evaluation process. Samples are generated via LLM-based synthetic sampling and human annotation.}
\label{fig:benchmark}
\end{figure*}

Large Language Models (LLMs) have become general-purpose tools for efficiently generating artifacts such as documents, code, and plans.
This generation process usually involves iterative cycles of generation and revision through conversations between users and LLMs. 
A problem here is that, when requesting revisions, users often find it difficult or burdensome to explicitly specify all parts of the artifact that are affected by the request.
Consequently, LLMs must instead identify the relevant parts and propagate the required revisions across the artifact to preserve consistency.

This problem has been studied extensively in LLM-based repository-level coding~\cite{jimenez2024swebench,codeplan,du-etal-2025-dependeval}, where LLMs must locate relevant files and propagate edits across codebase-wide dependencies.
More recently, similar propagation problems have been studied in LLM-based knowledge editing~\cite{rippleedit,dong-etal-2025-chainedit}, where factual edits are expected to propagate to related facts or logically connected knowledge, and in LLM-based document editing~\cite{ledger,editpropbench}, where local revisions may require updates to dependent document elements or claims to preserve consistency. 

However, these works focus on artifacts whose dependencies are largely explicit or statically analyzable: in coding, through call graphs, imports, and variable references; in knowledge editing, through pre-existing knowledge graphs; and in document editing, through references to sections, figures, citations, or claims.
In contrast, for conversationally generated artifacts to support practical tasks such as planning (Figure~\ref{fig:teaser}), dependencies among elements are often implicit and may be established through the surrounding conversational context (i.e., outside the artifact).
Revision propagation in this practical setting remains unexplored.

Motivated by this, this paper studies the ability of LLMs to propagate revisions across dependent elements in such conversationally generated artifacts\footnote{This paper focuses on JSON-formatted artifacts, as JSON is generally used as an output format in LLM systems.}.
Toward practical use, we also investigate how cost-effectively test-time compute improves the performance.
Specifically, we introduce RevPropBench, a new human-annotated benchmark for evaluating revision propagation in the new setting.
It consists of 30 development samples and 120 test samples across nine domains involving planning, record-keeping, and configuration, with three artifact sizes: 10, 50, and 100 JSON elements.
We then evaluate nine revision methods, including sequential reflection and parallel sampling variants, using gpt-oss-20b/120b, gpt-5.4-mini, qwen3.5-9b/27b/122b on the benchmark.

The results show that LLMs can propagate revisions in a single inference with an accuracy of 68.3--93\%, varying across models.
Among the test-time compute methods, 
selecting one of three parallel samples using either LLM-based or medoid-based selection is the most cost-effective, which improves accuracy by 2.2--9.7\%.

In summary, our contributions are threefold:
\begin{itemize}
    \item We propose a new benchmark for evaluating the ability of LLMs to propagate revisions across dependent elements in conversationally generated JSON artifacts (Section~\ref{sec:benchmark}).
    \item We comprehensively evaluate nine revision methods with six LLMs on the benchmark, providing practical guidance for cost-effective method selection (Section~\ref{sec:cost_analysis}).
    \item We release the benchmark instance, data sampling and annotation tool for reproducibility and future extension. 
\end{itemize}
\section{RevPropBench}
\label{sec:benchmark}
\begin{figure*}[t]
\centering
\begin{subfigure}[t]{0.32\textwidth}
  \centering
  \includegraphics[width=\linewidth]{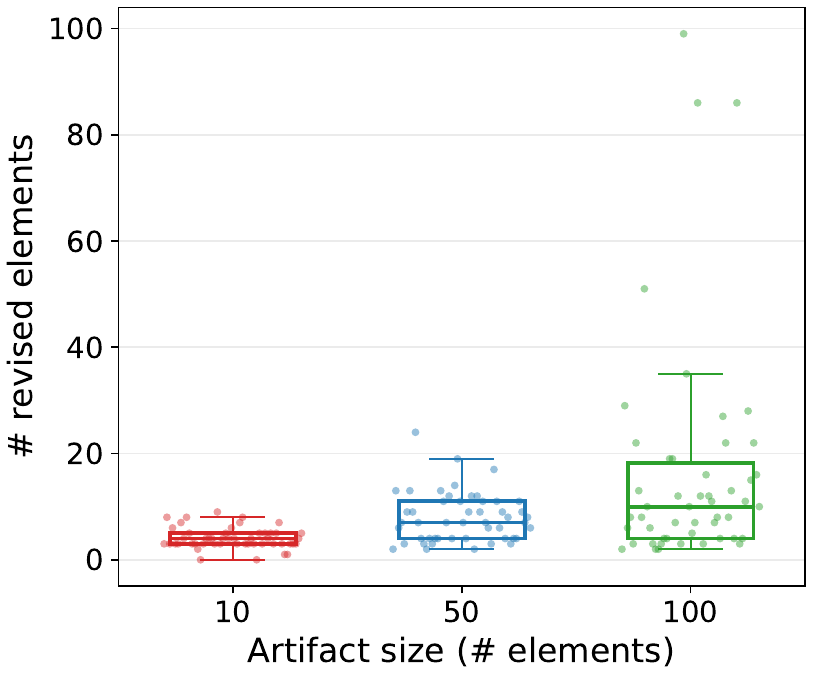}
  \caption{Propagation size by artifact size.}
  \label{fig:stats-tier}
\end{subfigure}
\hfill
\begin{subfigure}[t]{0.32\textwidth}
  \centering
  \includegraphics[width=\linewidth]{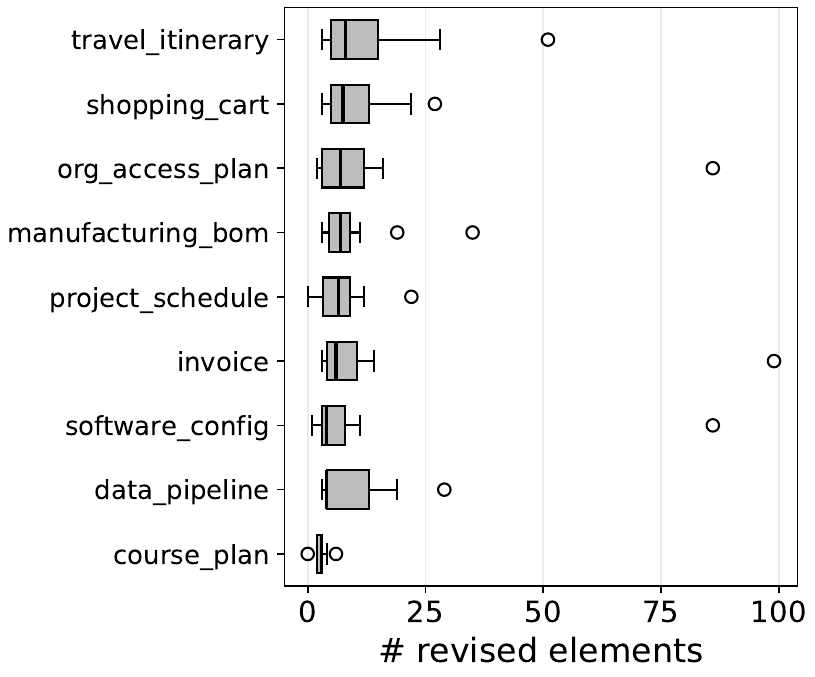}
  \caption{Propagation size by domain.}
  \label{fig:stats-domain}
\end{subfigure}
\hfill
\begin{subfigure}[t]{0.32\textwidth}
  \centering
  \includegraphics[width=\linewidth]{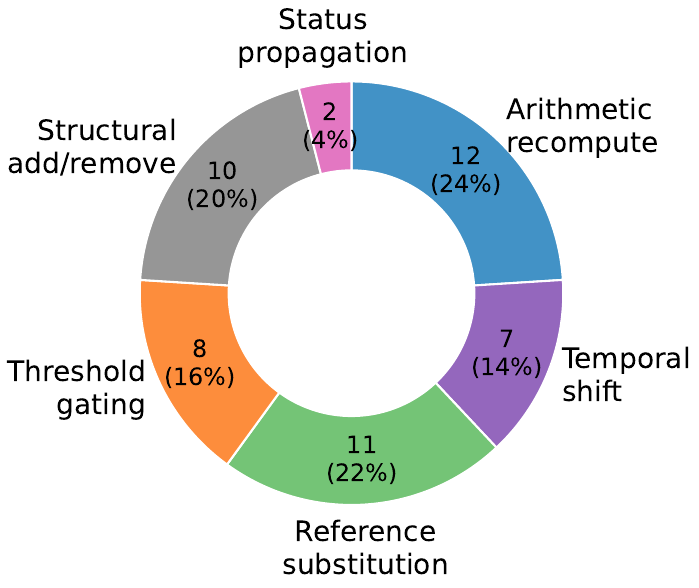}
  \caption{Propagation patterns.}
  \label{fig:stats-pattern}
\end{subfigure}
\caption{Statistics of RevPropBench ($n{=}150$ samples over $9$ domains and $50$ scenarios). }
\label{fig:stats}
\end{figure*}

Figure \ref{fig:benchmark} shows the overview of our proposed RevPropBench.
In the following, we describe the task definition, the benchmark construction process (data sampling, human annotation, and statistics), and the evaluation process (data split and metrics).

\subsection{Task definition}
The benchmark consists of two phases: a generation phase and a revision phase.
During the generation phase, an LLM incrementally adds elements to a JSON artifact through multiple turns of conversation with a user.
In the subsequent revision phase, the user provides a local revision request, and the LLM accordingly revises the relevant elements of the artifact by outputting JSON patches (RFC 6902).
The generation-phase conversation is prepared in advance, so the task is to revise the artifact in the revision phase.
Here, we evaluate whether the LLM can make exactly the necessary revisions to the artifact, i.e., whether the generated patches match the gold patches in both paths and values.

\subsection{Synthetic data sampling}
\label{sec:gengraph}
For each sample in the benchmark, we prepare a conversation history between an LLM and a user, including the generated JSON artifact, a local revision request from the user, and the corresponding gold patches.
The samples are synthetically generated using a strong LLM (e.g., GPT-5.5) for controllable data sampling.
We first create scenarios through LLM-assisted brainstorming, where each scenario specifies the conversation flow and the revision-propagation pattern.
Scenario design is guided by three criteria: covering diverse domains, covering diverse propagation patterns, and ensuring that the gold patches are deterministically defined for subsequent annotation.
We then use these scenarios to prompt the LLM to generate synthetic user--LLM conversation histories and local revision requests, similar to existing synthetic dialogue generation~\cite{ultrachat,baize}.

Each LLM turn includes patches that add at least one element to the artifact, and applying them sequentially across turns yields a single final artifact. 
The final artifact size is controlled by varying the maximum number of conversation turns: each conversation proceeds up to this limit but stops early once the target element count is reached.

\subsection{LLM-assisted human annotation}
For each generated sample, we annotate gold patches corresponding to the revision request.
We first use another strong LLM (e.g., Claude-Opus-4.8) to generate tentative gold patches for each sample. Human annotators then review and correct the patches through the GUI of our annotation tool, while consulting the LLM when necessary. See Appendix \ref{app:annotation} for details of the annotation tool.

A JSON patch (RFC 6902) is an ordered sequence of operations that modify a JSON artifact. 
Each operation is a triple $(\textit{op}, \textit{path}, \textit{value})$, where $\textit{op} \in \{\texttt{replace}, \texttt{add}, \texttt{remove}\}$ specifies the operation to apply, $\textit{path}$ specifies the target element, and $\textit{value}$ is the new content.
For free-form values that can be phrased in different ways, we define matchers that check for required keywords rather than exact string matches.
We also allow annotators to assign an optional flag to elements for which both editing and leaving unchanged are contextually valid.
Optional elements are treated as correct in either case and are counted as errors only when they are edited with an incorrect value.

\subsection{Metrics}
We evaluate LLMs by comparing their generated patches with the gold patches.
Our primary metric is the \emph{completion rate}: the proportion of samples for which the patched JSON artifact exactly matches the gold-patched artifact.
In failure analysis, we also report three finer-grained metrics: \emph{miss} for omitted necessary edits, \emph{over edit} for unnecessary edits, and \emph{wrong value} for incorrect values in necessary edits.

\subsection{Benchmark instance}
In this paper, we create 50 scenarios across nine practical domains, including travel itineraries, invoices, shopping carts, project schedules, course plans, data pipelines, software deployment configurations, organizational access plans, and manufacturing BOMs. For each scenario, we generate three samples with JSON artifacts containing 10 (small), 50 (medium), and 100 (large) elements, resulting in 50 × 3 = 150 samples in total.

Figure~\ref{fig:stats} summarizes the statistics of the instantiated benchmark.
The number of propagated revisions increases as the number of artifact elements grows (Figure~\ref{fig:stats-tier}), while the distribution of characteristic propagation patterns across scenarios illustrates the diversity of revision-propagation cases covered by the benchmark (Figure~\ref{fig:stats-pattern}).

\section{Evaluation Settings}
\label{sec:setup}

\paragraph{Data split}
We divide the 150 samples into a development split for prompt tuning and a test split for evaluation.
The split is performed at the scenario level, with all three size variants of each scenario assigned to the same split.
We assign 10 of the 50 scenarios (i.e., 30 samples) to development via random sampling with seed 42, balanced by domain, propagation size, and propagation pattern.
The remaining 120 samples form the test split, on which all reported metrics are computed.

\paragraph{Evaluated revision methods}
We evaluate nine methods, covering single-pass baselines and test-time compute methods, including sequential reflection and parallel sampling variants:
\begin{itemize}
    \item \textbf{Baselines} generate a patch in a single inference, with the final JSON artifact (\textbf{\textsc{j}}), the conversation history (\textbf{\textsc{h}}), or both (\textbf{\textsc{j+h}}) provided as auxiliary context.
    \item \textbf{Sequential reflection (\textsc{Reflect})} starts from the patch generated with \textsc{j+h} and iteratively refines it through reflection.
    \item \textbf{Parallel sampling with rule-based selection}, similar to self-consistency~\cite{wang2023selfconsistency}, first generates multiple patches in parallel using \textsc{j+h}, and then generates the final patch from them according to predefined rules. \textbf{\textsc{or}}, \textbf{\textsc{and}}, and \textbf{\textsc{maj}} update a leaf element when any candidate changes it, when all candidates agree on the same new value, and when a strict majority agrees on the same new value, respectively. \textbf{\textsc{med}} (medoid), inspired by minimum Bayes-risk decoding~\cite{mbr}, selects the candidate whose resulting artifact has the smallest mean leaf-level disagreement with the others. 
    See Appendix~\ref{app:methods} for more details.
    \item \textbf{Parallel sampling with LLM-based selection (\textsc{Select})} first generates multiple patches in parallel using \textsc{j+h}, and then selects one of them as the final patch using the same LLM. 
\end{itemize}

\paragraph{Evaluated LLMs}
We use six representative LLMs spanning two model families and three different parameter scales: gpt-oss-20b/120b~\cite{openai2025gptoss120bgptoss20bmodel}, gpt-5.4-mini~\cite{gpt-5.4-mini}, and qwen3.5-9b/27b/122b-a10b~\cite{qwen3.5}.
Reasoning is enabled for all models, and the effort level is set to medium for the GPT models.

\paragraph{Hyperparameters}
We set the temperature to 0.6 for all models that support it, except for gpt-5.4-mini, and set the maximum output context length to 32K tokens. Local models are deployed using vLLM~\cite{vllm} on four NVIDIA A100 (80GB) GPUs.
We run five evaluations for each sample with seeds $s=0,42,84,126,168$.
\textsc{Reflect} performs four reflection iterations, i.e., five LLM calls in total, using the same seed $s$.
For parallel sampling in each run, we sample five outputs with seeds $s,s+1,\dots,s+4$.
Note that \textsc{Select} samples four outputs and uses seed $s$ for the additional LLM-based selection call, and that LiteLLM caching~\cite{litellm} ensures identical outputs for the same seed and input.
\section{Evaluation Results}
\label{sec:experiments}

\begin{figure*}[t]
\centering
\includegraphics[width=\linewidth]{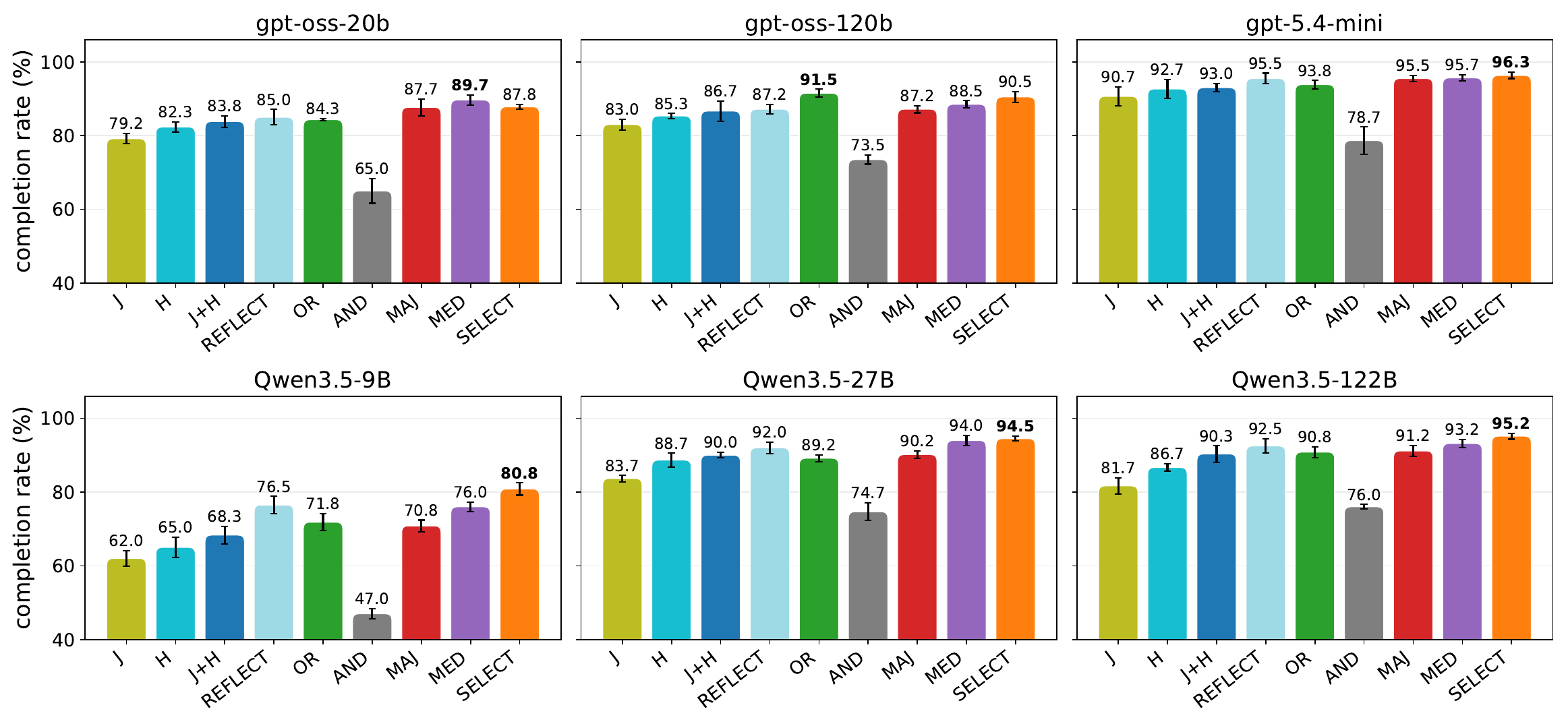}
\caption{Completion rate of the nine revision methods on the 120-sample test split, for each of the six models. Bars are the mean over five different seeds and error bars the standard deviation. The test-time compute methods (\textsc{Reflect}, \textsc{or}, \textsc{and}, \textsc{maj}, \textsc{med}, and \textsc{Select}) are evaluated with five LLM calls.}
\label{fig:main_result}
\end{figure*}

\subsection{Performance}
Figure~\ref{fig:main_result} reports the completion rate of the nine revision methods across the six LLMs on the test split, averaged over five runs.

\paragraph{Importance of conversation history}
Across every model, the baselines follow a consistent performance ordering, \textsc{j} < \textsc{h} < \textsc{j+h} (e.g., 90.7\% < 92.7\% < 93.0\% for gpt-5.4-mini and 81.7\% < 86.7\% < 90.3\% for qwen3.5-122b).
The improvement from \textsc{j} to \textsc{h} suggests that the conversation history contains important contextual information about the elements and their dependencies that are not fully recoverable from the final artifact alone.
This reflects our benchmark design for conversationally generated artifacts.
The further improvement from \textsc{h} to \textsc{j+h} indicates that, although the conversation history already contains the information needed to reconstruct the final artifact through patches, explicitly re-providing the final artifact helps LLMs avoid path errors and missed revisions when generating JSON patches.

\paragraph{Model comparison}
Performance varies widely across models. With the strongest baseline (\textsc{j+h}), for example, completion rates span 68.3--93.0\%. 
Generally, they tend to be higher for models with larger scale, i.e., qwen3.5-9b < 27b < 122b and gpt-oss-20b < 120b < gpt-5.4-mini.

\paragraph{Method comparison}
Among the test-time compute methods, \textsc{Select} yields the most consistent improvement over \textsc{j+h} ($+3.3\textrm{--}12.5\%$), achieving the highest completion rate on four of the six models and the second-highest on the remaining two.
\textsc{med} provides the second-most consistent improvement ($+1.8\textrm{--}7.7\%$), achieving the highest completion rate on one model and the second-highest on three models.
The rule-based merging methods are far less reliable: \textsc{and} falls well below \textsc{j+h} on every model ($-13.2\textrm{--}21.3\%$), since demanding unanimous agreement discards correct edits that only some samples find, while \textsc{or} and \textsc{maj} are competitive with \textsc{Select} and \textsc{med} on a few models but are not consistent across models ($-0.8\textrm{--}{+}4.8\%$).
\textsc{Reflect} yields small, consistent gains ($+0.5\textrm{--}8.2\%$) but rarely matches \textsc{Select} or \textsc{med}.

\begin{figure}[t]
\centering
\includegraphics[width=0.95\linewidth]{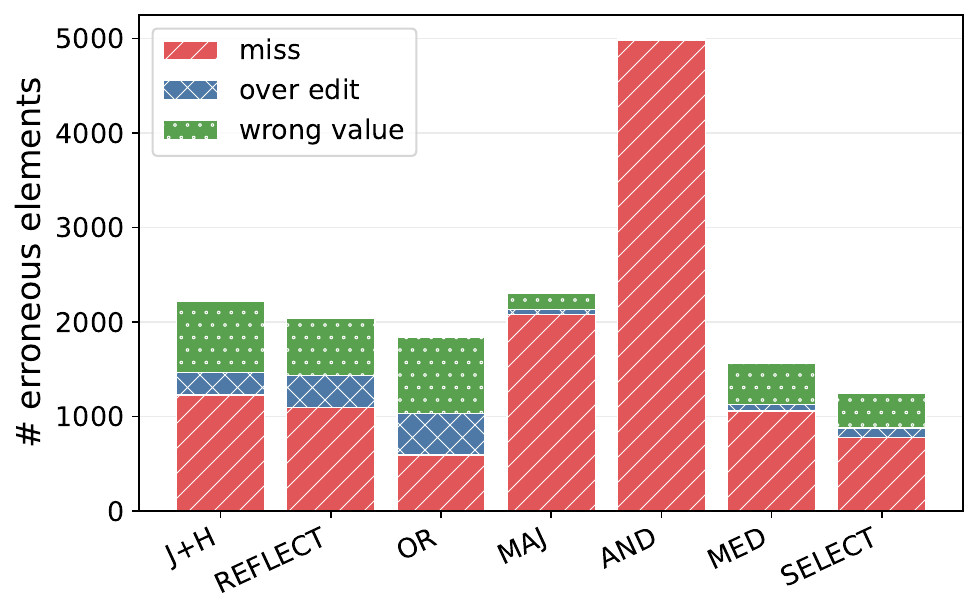}
\caption{
Element-level failure composition by method, pooled over the six models (i.e., 3600 samples)
} 
\label{fig:failure}
\end{figure}

\begin{figure*}[t]
\centering
\includegraphics[width=\linewidth]{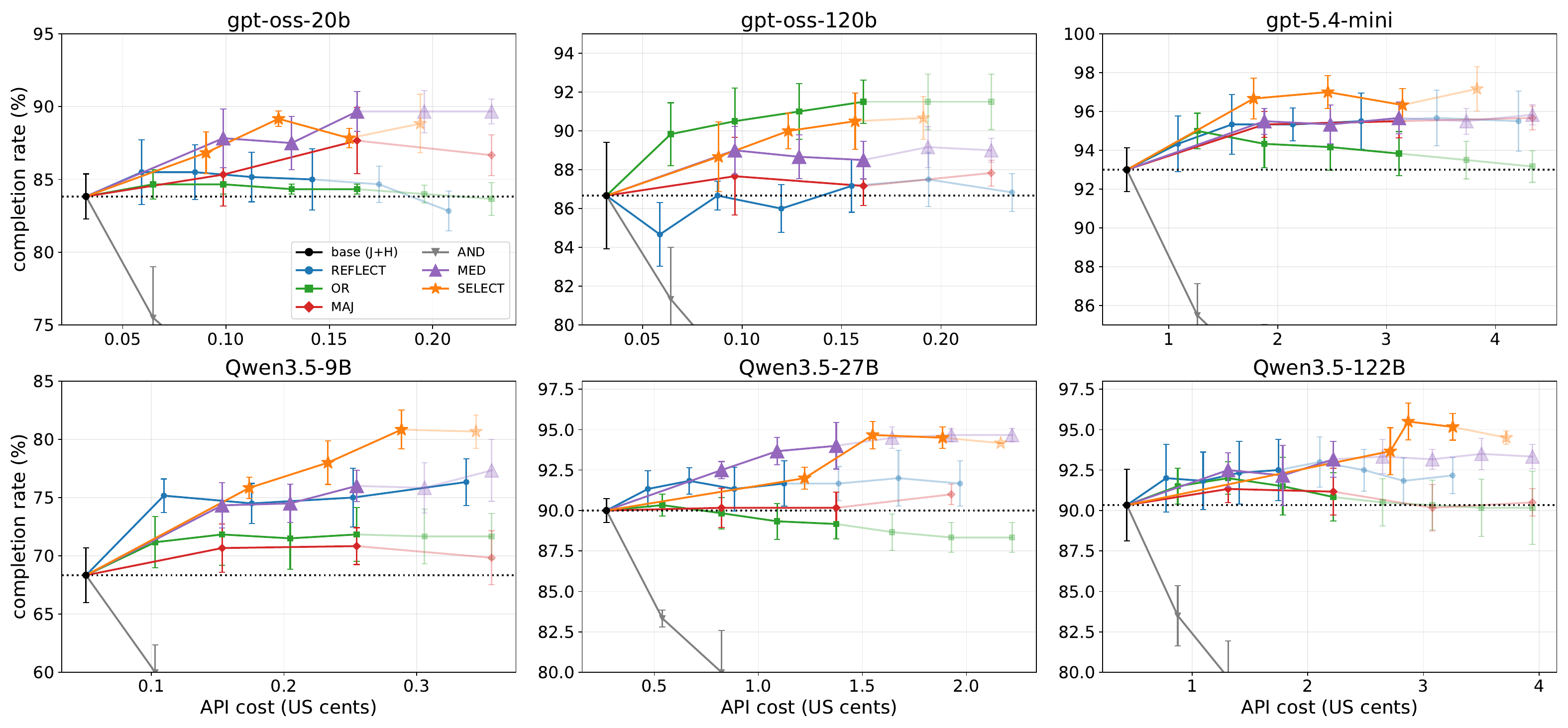}
\caption{
Completion rate versus API cost as the number of LLM calls increases.
Solid markers show points up to five LLM calls, while faded markers show additional points for analyzing performance convergence.
}
\label{fig:cost_curve}
\end{figure*}

\paragraph{Failure analysis}
We group failures into \emph{miss}, \emph{over edit}, and \emph{wrong value}. Figure~\ref{fig:failure} summarizes their composition by method. 
For most methods, \emph{miss} accounts for the majority of errors: models tend to propagate too little rather than too much.
\textsc{Reflect} reduces misses through iterative revision, but the reduction is limited.
\textsc{and} fails almost only by miss because its strict unanimity requirement discards even correct edits if any candidate differs, whereas \textsc{or} introduces many over edits by accepting any proposed edit.
\textsc{maj} lies between \textsc{and} and \textsc{or}.
\textsc{med} and \textsc{Select} produce the fewest failures by returning one whole candidate: \textsc{med} reduces outlier over-edits by choosing the most typical sample but recovers fewer misses, whereas \textsc{Select}'s reasoning tends to target the most complete candidate, reducing misses the most but not over edits.

\subsection{Cost analysis}
\label{sec:cost_analysis}
Figure~\ref{fig:cost_curve} plots completion rate against API cost\footnote{For local models, we estimate API cost using the OpenRouter pricing table.}, with each curve showing how performance and cost change for each model and method as the number of LLM calls increases. 
Solid markers show points up to the five LLM calls set as a hyperparameter, while faded markers show additional points with more calls for analyzing cost-performance scaling.
Note that we plot \textsc{maj} only at odd call counts, and \textsc{Select} from three calls onward (2 sampling + 1 selection calls).

\paragraph{Cost comparison under fixed LLM calls}
At five LLM calls, corresponding to the rightmost solid markers, GPT models consume similar amounts of API cost across all methods, with costs increasing roughly in proportion to the number of calls from \textsc{j+h}.
In contrast, for Qwen models, \textsc{Select} tends to incur higher costs than the other methods, increasing the cost by $5.7\textrm{--}7.5\times$ over \textsc{j+h}. This difference is mainly due to the larger number of reasoning tokens generated in \textsc{Select}.

\paragraph{Performance comparison under fixed cost}
To compare performance more fairly with respect to cost, we next compare methods after aligning their costs. 
Specifically, we extend the number of calls for the other methods until their costs become comparable to, or higher than, that of \textsc{Select}.
Even under this alignment, the relative performance ordering among methods remains largely unchanged, indicating that \textsc{Select}'s advantage is not merely due to higher cost.

\paragraph{Cost-performance scaling}
Finally, we analyze how performance improves relative to the additional cost based on the full curves, including both solid and faded points.
Figure~\ref{fig:cost_curve} shows that performance largely saturates around four to five LLM calls for most methods and models.
In this post-hoc analysis, we find that, on average across models, the highest cost-effectiveness $(\frac{\mathrm{gain}}{\mathrm{cost} / \mathrm{cost}_{\textsc{j+h}}-1})$ is achieved by either \textsc{Select} with three or four LLM calls, or \textsc{med} with three LLM calls.

\paragraph{Guidelines for cost-effective test-time compute}
Based on the above analyses and absolute values, we recommend \textsc{Select} with four LLM calls as a practical choice for cost-effectively improving revision propagation through test-time compute.
On the other hand, it requires two sequential stages of sampling and selection, whereas \textsc{med} only requires a single parallel sampling stage. Thus, when inference latency is also considered, \textsc{med} with three LLM calls is an alternative choice, offering competitive performance with \textsc{Select} at lower latency.


\section{Related Work}
\paragraph{Revision Propagation}
Revision propagation refers to the problem of applying a local edit and updating other parts that depend on it to preserve global consistency.
This problem has been studied actively in repository-level code editing~\cite{jimenez2024swebench,codeplan,du-etal-2025-dependeval}, where, given a task, LLMs must identify the required edits and propagate them across all dependent parts of the repository. 
Recent work has shown that explicitly extracted repository graphs can improve such repository-level coding~\cite{graphcoder,codexgraph,repograph,coret,cgm}. 

Knowledge editing studies how to update factual knowledge encoded in LLMs. In this setting, revision propagation is studied as a ripple effect: after editing one fact, the model should also update related facts that are affected by the edit.
RippleEdits~\cite{rippleedit} evaluates this by querying the edited model about both the target fact and related facts that should change or remain unchanged after the edit.
ChainEdit~\cite{dong-etal-2025-chainedit} improves ripple-effect propagation by extracting logical rules from knowledge graphs and using them with LLM reasoning to update chains of logically connected facts.

Document editing studies how to revise long-form documents while preserving consistency across dependent textual elements. 
In this setting, revision propagation arises when a local edit affects other claims, references, or descriptions elsewhere in the document.
EditPropBench~\cite{editpropbench} evaluates factual edit propagation in scientific manuscripts by testing whether LLMs update not only the directly edited sentence but also dependent manuscript claims that should change, while leaving unrelated text unchanged. 
LEDGER~\cite{ledger} improves agentic document editing by constructing a dependency graph from both document structure and LLM-inferred semantic dependencies, and using graph-guided retrieval to select only the necessary context for each edit.

Our work studies revision propagation in a different setting.
Unlike prior settings that can rely on pre-existing dependency sources such as repository graphs, knowledge graphs, document structure, or explicit references, our setting often lacks such dependency information in advance. 
Additionally, a distinctive feature of our setting is that dependencies may be established implicitly during the multi-turn conversation that generated the artifact, outside the artifact itself.

\paragraph{JSON artifacts in LLM-based systems}
JSON is a standard output format for LLM systems, used for tool calls, agent actions, or structured outputs for system-side data processing~\cite{json,tool}.
Therefore, considering both practical utility and the ease of automated evaluation, this paper adopts JSON artifacts as the output format for LLMs during conversations.
Similar to our work, \citet{duanis-etal-2025-json} study the performance of LLMs in generating JSON patches. However, their focus is on the accuracy of modifying the JSON artifact itself, and their setting does not consider the conversational process that generated it. 
In addition, our work has a different objective: to identify cost-effective test-time compute methods for revision propagation.

\paragraph{Test-time compute}
Test-time compute improves model outputs by increasing inference-time computation through reasoning, sampling, or search.
Representative parallel sampling approaches include self-consistency~\cite{wang2023selfconsistency}, which selects the most consistent answer from sampled reasoning paths; best-of-N~\cite{cobbe2021trainingverifierssolvemath}, which chooses among generated candidates using a learned verifier or reward model.
Other approaches iteratively improve outputs in a sequential manner through reflection or self-feedback, such as Self-Refine~\cite{madaan2023selfrefine} and Reflexion~\cite{shinn2023reflexion}.
In this paper, we study whether test-time compute remains effective in our new problem setting and investigate how many samples provide the best cost–performance trade-off.
\section{Conclusion}
In this paper, we proposed a new benchmark for evaluating the ability of LLMs to propagate revisions across JSON artifacts generated through conversation. We then studied how test-time compute improves the performance, demonstrating that selecting one of three parallel samples using either LLM-based or medoid-based selection is the most cost-effective, and this improves accuracy by 2.2--9.7\%, compared to a single inference.
Our findings provide practical insights into how we can improve revision reliability by test-time compute while balancing performance gains and inference cost in the new problem setting.

\section*{Limitations}
\paragraph{Scope} This work does not generalize to the revision propagation problems studied in existing work on repository-level code editing, knowledge editing, or document editing.
For these problems, prior work has already shown that performance can be improved by explicitly providing dependency information that is largely available in advance, such as statically analyzable code dependency graphs, knowledge graphs, and chapter or reference structures.
On the other hand, this work focuses on a distinct setting in which dependencies are unavailable in advance and may be established by the conversation outside the artifact.

\paragraph{Gap from real-world conversations} Although we created scenarios that approximate real-world use cases, the conversations are still synthetically generated by an LLM based on those scenarios to enable controllable evaluation. We do not use actual human–LLM interaction data. 
Furthermore, to enable reliable automatic evaluation, we control the dependencies established during the conversation to be deterministic. As a result, this work does not cover revision propagation in ambiguous cases that can arise in real conversations, where people may disagree on whether a particular element should be jointly revised.

\paragraph{Data coverage and potential biases}
The benchmark consists of 150 samples covering nine practical domains, six propagation patterns, and three artifact sizes. Although it covers a broad range of representative patterns, it does not exhaustively cover all possible patterns.
In terms of preference biases, since we do not use an LLM-as-a-judge, there is no concern regarding the use of models from the same GPT family for both data generation and evaluation. On the other hand, because the conversations were generated using GPT-5.5, the resulting samples may be easier for GPT models. In this paper, the scenarios were created by humans with Claude-Opus-4.8, and the conversations were constructed from meta-level instructions such that task overlap between data generation and evaluation is small, thereby mitigating such a bias. 

\paragraph{Performance saturation}
Currently, the baseline completion rates of six representative models range from 68.3\% to 93\%, allowing meaningful comparisons across models and methods. However, gpt-5.4-mini already achieves a completion rate of 93\%, raising concerns that, as future models become more capable, performance may saturate and make such comparisons increasingly difficult.
To address this limitation, this work provides not only the benchmark instances but also a tool that supports the entire benchmark construction process, from sample generation to annotation. Using this tool, the benchmark will need to be continuously updated to increase its difficulty as model capabilities improve.

\section*{Ethical Considerations}
Since our benchmark is synthetic, it does not raise direct privacy concerns related to personal information.
Moreover, during scenario construction and after data sampling, we carefully reviewed the scenarios and generated samples for ethically problematic content. We confirmed that none of the samples contained such content.
\bibliography{custom}

@inproceedings{ledger,
    title = "{LEDGER}: Scaling Agentic Document Editing with Dependency-aware Graph Retrieval",
    author = "Wang, Hang  and
      Garg, Utkarsh  and
      Davari, Reza  and
      Jiao, Huitian  and
      Cheng, Hao  and
      Peng, Baolin  and
      Chen, Si-Qing  and
      Ge, Tao",
    editor = "Liakata, Maria  and
      Moreira, Viviane P.  and
      Zhang, Jiajun  and
      Jurgens, David",
    booktitle = "Findings of the {A}ssociation for {C}omputational {L}inguistics: {ACL} 2026",
    month = jul,
    year = "2026",
    address = "San Diego, California, United States",
    publisher = "Association for Computational Linguistics",
    url = "https://aclanthology.org/2026.findings-acl.515/",
    doi = "10.18653/v1/2026.findings-acl.515",
    pages = "10614--10644",
    ISBN = "979-8-89176-395-1"
}

@misc{editpropbench,
      title={{EditPropBench}: Measuring Factual Edit Propagation in Scientific Manuscripts}, 
      author={Garvin Kruthof},
      year={2026},
      eprint={2605.02083},
      archivePrefix={arXiv},
      primaryClass={cs.CL},
      url={https://arxiv.org/abs/2605.02083}, 
}

@article{rippleedit,
    title = "Evaluating the Ripple Effects of Knowledge Editing in Language Models",
    author = "Cohen, Roi  and
      Biran, Eden  and
      Yoran, Ori  and
      Globerson, Amir  and
      Geva, Mor",
    journal = "Transactions of the Association for Computational Linguistics",
    volume = "12",
    year = "2024",
    address = "Cambridge, MA",
    publisher = "MIT Press",
    url = "https://aclanthology.org/2024.tacl-1.16/",
    doi = "10.1162/tacl_a_00644",
    pages = "283--298"
}

@inproceedings{dong-etal-2025-chainedit,
    title = "{C}hain{E}dit: Propagating Ripple Effects in {LLM} Knowledge Editing through Logical Rule-Guided Chains",
    author = "Dong, Zilu  and
      Shen, Xiangqing  and
      Yang, Zinong  and
      Xia, Rui",
    editor = "Che, Wanxiang  and
      Nabende, Joyce  and
      Shutova, Ekaterina  and
      Pilehvar, Mohammad Taher",
    booktitle = "Proceedings of the 63rd Annual Meeting of the Association for Computational Linguistics (Volume 1: Long Papers)",
    month = jul,
    year = "2025",
    address = "Vienna, Austria",
    publisher = "Association for Computational Linguistics",
    url = "https://aclanthology.org/2025.acl-long.665/",
    doi = "10.18653/v1/2025.acl-long.665",
    pages = "13558--13571",
    ISBN = "979-8-89176-251-0"
}

@inproceedings{du-etal-2025-dependeval,
    title = "{D}epend{E}val: Benchmarking {LLM}s for Repository Dependency Understanding",
    author = "Du, Junjia  and
      Liu, Yadi  and
      Guo, Hongcheng  and
      Wang, Jiawei  and
      Huang, Haojian  and
      Ni, Yunyi  and
      Li, Zhoujun",
    editor = "Che, Wanxiang  and
      Nabende, Joyce  and
      Shutova, Ekaterina  and
      Pilehvar, Mohammad Taher",
    booktitle = "Findings of the Association for Computational Linguistics: ACL 2025",
    month = jul,
    year = "2025",
    address = "Vienna, Austria",
    publisher = "Association for Computational Linguistics",
    url = "https://aclanthology.org/2025.findings-acl.373/",
    doi = "10.18653/v1/2025.findings-acl.373",
    pages = "7150--7179",
    ISBN = "979-8-89176-256-5"
}

@article{codeplan,
author = {Bairi, Ramakrishna and Sonwane, Atharv and Kanade, Aditya and C., Vageesh D. and Iyer, Arun and Parthasarathy, Suresh and Rajamani, Sriram and Ashok, B. and Shet, Shashank},
title = {{CodePlan}: Repository-Level Coding using {LLM}s and Planning},
year = {2024},
issue_date = {July 2024},
publisher = {Association for Computing Machinery},
address = {New York, NY, USA},
volume = {1},
number = {FSE},
url = {https://doi.org/10.1145/3643757},
doi = {10.1145/3643757},
journal = {Proc. ACM Softw. Eng.},
month = jul,
articleno = {31},
numpages = {24},
pages={675--698}
}

@inproceedings{
jimenez2024swebench,
title={{SWE}-bench: Can Language Models Resolve Real-world {Github} Issues?},
author={Carlos E Jimenez and John Yang and Alexander Wettig and Shunyu Yao and Kexin Pei and Ofir Press and Karthik R Narasimhan},
booktitle={The Twelfth International Conference on Learning Representations},
year={2024},
url={https://openreview.net/forum?id=VTF8yNQM66}
}

@inproceedings{ultrachat,
    title = "Enhancing Chat Language Models by Scaling High-quality Instructional Conversations",
    author = "Ding, Ning  and
      Chen, Yulin  and
      Xu, Bokai  and
      Qin, Yujia  and
      Hu, Shengding  and
      Liu, Zhiyuan  and
      Sun, Maosong  and
      Zhou, Bowen",
    editor = "Bouamor, Houda  and
      Pino, Juan  and
      Bali, Kalika",
    booktitle = "Proceedings of the 2023 Conference on Empirical Methods in Natural Language Processing",
    month = dec,
    year = "2023",
    address = "Singapore",
    publisher = "Association for Computational Linguistics",
    url = "https://aclanthology.org/2023.emnlp-main.183/",
    doi = "10.18653/v1/2023.emnlp-main.183",
    pages = "3029--3051"
}

@inproceedings{baize,
    title = "Baize: An Open-Source Chat Model with Parameter-Efficient Tuning on Self-Chat Data",
    author = "Xu, Canwen  and
      Guo, Daya  and
      Duan, Nan  and
      McAuley, Julian",
    editor = "Bouamor, Houda  and
      Pino, Juan  and
      Bali, Kalika",
    booktitle = "Proceedings of the 2023 Conference on Empirical Methods in Natural Language Processing",
    month = dec,
    year = "2023",
    address = "Singapore",
    publisher = "Association for Computational Linguistics",
    url = "https://aclanthology.org/2023.emnlp-main.385/",
    doi = "10.18653/v1/2023.emnlp-main.385",
    pages = "6268--6278"
}

@misc{openai2025gptoss120bgptoss20bmodel,
      title={gpt-oss-120b \& gpt-oss-20b Model Card}, 
      author={OpenAI and : and Sandhini Agarwal and Lama Ahmad and Jason Ai and Sam Altman and Andy Applebaum and Edwin Arbus and Rahul K. Arora and Yu Bai and Bowen Baker and Haiming Bao and Boaz Barak and Ally Bennett and Tyler Bertao and Nivedita Brett and Eugene Brevdo and Greg Brockman and Sebastien Bubeck and Che Chang and Kai Chen and Mark Chen and Enoch Cheung and Aidan Clark and Dan Cook and Marat Dukhan and Casey Dvorak and Kevin Fives and Vlad Fomenko and Timur Garipov and Kristian Georgiev and Mia Glaese and Tarun Gogineni and Adam Goucher and Lukas Gross and Katia Gil Guzman and John Hallman and Jackie Hehir and Johannes Heidecke and Alec Helyar and Haitang Hu and Romain Huet and Jacob Huh and Saachi Jain and Zach Johnson and Chris Koch and Irina Kofman and Dominik Kundel and Jason Kwon and Volodymyr Kyrylov and Elaine Ya Le and Guillaume Leclerc and James Park Lennon and Scott Lessans and Mario Lezcano-Casado and Yuanzhi Li and Zhuohan Li and Ji Lin and Jordan Liss and Lily and Liu and Jiancheng Liu and Kevin Lu and Chris Lu and Zoran Martinovic and Lindsay McCallum and Josh McGrath and Scott McKinney and Aidan McLaughlin and Song Mei and Steve Mostovoy and Tong Mu and Gideon Myles and Alexander Neitz and Alex Nichol and Jakub Pachocki and Alex Paino and Dana Palmie and Ashley Pantuliano and Giambattista Parascandolo and Jongsoo Park and Leher Pathak and Carolina Paz and Ludovic Peran and Dmitry Pimenov and Michelle Pokrass and Elizabeth Proehl and Huida Qiu and Gaby Raila and Filippo Raso and Hongyu Ren and Kimmy Richardson and David Robinson and Bob Rotsted and Hadi Salman and Suvansh Sanjeev and Max Schwarzer and D. Sculley and Harshit Sikchi and Kendal Simon and Karan Singhal and Yang Song and Dane Stuckey and Zhiqing Sun and Philippe Tillet and Sam Toizer and Foivos Tsimpourlas and Nikhil Vyas and Eric Wallace and Xin Wang and Miles Wang and Olivia Watkins and Kevin Weil and Amy Wendling and Kevin Whinnery and Cedric Whitney and Hannah Wong and Lin Yang and Yu Yang and Michihiro Yasunaga and Kristen Ying and Wojciech Zaremba and Wenting Zhan and Cyril Zhang and Brian Zhang and Eddie Zhang and Shengjia Zhao},
      year={2025},
      eprint={2508.10925},
      archivePrefix={arXiv},
      primaryClass={cs.CL},
      url={https://arxiv.org/abs/2508.10925}, 
}

@misc{qwen3.5,
    title  = {{Qwen3.5}: Towards Native Multimodal Agents},
    author = {{Qwen Team}},
    month  = {February},
    year   = {2026},
    url    = {https://qwen.ai/blog?id=qwen3.5},
    note   = {Accessed 2026-06-16}
}

@misc{gpt-5.4-mini,
    title  = {Introducing {GPT‑5.4} mini and nano},
    author = {{OpenAI}},
    month  = {March},
    year   = {2026},
    url    = {https://openai.com/index/introducing-gpt-5-4-mini-and-nano/},
    note   = {Accessed 2026-06-16}
}

@inproceedings{vllm,
author = {Kwon, Woosuk and Li, Zhuohan and Zhuang, Siyuan and Sheng, Ying and Zheng, Lianmin and Yu, Cody Hao and Gonzalez, Joseph and Zhang, Hao and Stoica, Ion},
title = {Efficient Memory Management for Large Language Model Serving with {PagedAttention}},
year = {2023},
isbn = {9798400702297},
publisher = {Association for Computing Machinery},
address = {New York, NY, USA},
url = {https://doi.org/10.1145/3600006.3613165},
doi = {10.1145/3600006.3613165},
booktitle = {Proceedings of the 29th Symposium on Operating Systems Principles},
pages = {611–626},
numpages = {16},
location = {Koblenz, Germany},
series = {SOSP '23}
}

@inproceedings{
wang2023selfconsistency,
title={Self-Consistency Improves Chain of Thought Reasoning in Language Models},
author={Xuezhi Wang and Jason Wei and Dale Schuurmans and Quoc V Le and Ed H. Chi and Sharan Narang and Aakanksha Chowdhery and Denny Zhou},
booktitle={The Eleventh International Conference on Learning Representations },
year={2023},
url={https://openreview.net/forum?id=1PL1NIMMrw}
}

@inproceedings{mbr,
    title = "Sampling-Based Approximations to Minimum {B}ayes Risk Decoding for Neural Machine Translation",
    author = "Eikema, Bryan  and
      Aziz, Wilker",
    editor = "Goldberg, Yoav  and
      Kozareva, Zornitsa  and
      Zhang, Yue",
    booktitle = "Proceedings of the 2022 Conference on Empirical Methods in Natural Language Processing",
    month = dec,
    year = "2022",
    address = "Abu Dhabi, United Arab Emirates",
    publisher = "Association for Computational Linguistics",
    url = "https://aclanthology.org/2022.emnlp-main.754/",
    doi = "10.18653/v1/2022.emnlp-main.754",
    pages = "10978--10993"
}

@inproceedings{
repograph,
title={{RepoGraph}: Enhancing {AI} Software Engineering with Repository-level Code Graph},
author={Siru Ouyang and Wenhao Yu and Kaixin Ma and Zilin Xiao and Zhihan Zhang and Mengzhao Jia and Jiawei Han and Hongming Zhang and Dong Yu},
booktitle={The Thirteenth International Conference on Learning Representations},
year={2025},
url={https://openreview.net/forum?id=dw9VUsSHGB}
}

@inproceedings{codexgraph,
    title = "{C}odex{G}raph: Bridging Large Language Models and Code Repositories via Code Graph Databases",
    author = "Liu, Xiangyan  and
      Lan, Bo  and
      Hu, Zhiyuan  and
      Liu, Yang  and
      Zhang, Zhicheng  and
      Wang, Fei  and
      Shieh, Michael Qizhe  and
      Zhou, Wenmeng",
    editor = "Chiruzzo, Luis  and
      Ritter, Alan  and
      Wang, Lu",
    booktitle = "Proceedings of the 2025 Conference of the Nations of the Americas Chapter of the Association for Computational Linguistics: Human Language Technologies (Volume 1: Long Papers)",
    month = apr,
    year = "2025",
    address = "Albuquerque, New Mexico",
    publisher = "Association for Computational Linguistics",
    url = "https://aclanthology.org/2025.naacl-long.7/",
    doi = "10.18653/v1/2025.naacl-long.7",
    pages = "142--160",
    ISBN = "979-8-89176-189-6"
}

@inproceedings{cgm,
 author = {Tao, Hongyuan and Zhang, Ying and Tang, Zhenhao and Peng, Hongen and Zhu, Xukun and Liu, Bingchang and Yang, Yingguang and Zhang, Ziyin and Xu, Zhaogui and Zhang, Haipeng and Zhu, Linchao and Wang, Rui and Yu, Hang and Li, Jianguo and Di, Peng},
 booktitle = {Advances in Neural Information Processing Systems},
 doi = {10.52202/085713-0537},
 editor = {D. Belgrave and C. Zhang and H. Lin and R. Pascanu and P. Koniusz and M. Ghassemi and N. Chen},
 pages = {15869--15909},
 publisher = {Curran Associates, Inc.},
 title = {{Code Graph Model} ({CGM}): A Graph-Integrated Large Language Model for Repository-Level Software Engineering Tasks},
 url = {https://proceedings.neurips.cc/paper_files/paper/2025/file/178ae4ba29022eb7bf509c2e27bc8ab8-Paper-Conference.pdf},
 volume = {38, Main Conference},
 year = {2025}
}

@inproceedings{coret,
    title = "{C}o{R}et: Improved Retriever for Code Editing",
    author = "Fehr, Fabio James  and
      Teja S, Prabhu  and
      Franceschi, Luca  and
      Zappella, Giovanni",
    editor = "Che, Wanxiang  and
      Nabende, Joyce  and
      Shutova, Ekaterina  and
      Pilehvar, Mohammad Taher",
    booktitle = "Proceedings of the 63rd Annual Meeting of the Association for Computational Linguistics (Volume 2: Short Papers)",
    month = jul,
    year = "2025",
    address = "Vienna, Austria",
    publisher = "Association for Computational Linguistics",
    url = "https://aclanthology.org/2025.acl-short.62/",
    doi = "10.18653/v1/2025.acl-short.62",
    pages = "775--789",
    ISBN = "979-8-89176-252-7"
}

@inproceedings{graphcoder,
author = {Liu, Wei and Yu, Ailun and Zan, Daoguang and Shen, Bo and Zhang, Wei and Zhao, Haiyan and Jin, Zhi and Wang, Qianxiang},
title = {{GraphCoder}: Enhancing Repository-Level Code Completion via Coarse-to-fine Retrieval Based on Code Context Graph},
year = {2024},
isbn = {9798400712487},
publisher = {Association for Computing Machinery},
address = {New York, NY, USA},
url = {https://doi.org/10.1145/3691620.3695054},
doi = {10.1145/3691620.3695054},
booktitle = {Proceedings of the 39th IEEE/ACM International Conference on Automated Software Engineering},
pages = {570–581},
numpages = {12},
location = {Sacramento, CA, USA},
series = {ASE '24}
}

@inproceedings{duanis-etal-2025-json,
    title = "{JSON} Whisperer: Efficient {JSON} Editing with {LLM}s",
    author = "Duanis, Sarel  and
      Greenstein-Messica, Asnat  and
      Habba, Eliya",
    editor = "Potdar, Saloni  and
      Rojas-Barahona, Lina  and
      Montella, Sebastien",
    booktitle = "Proceedings of the 2025 Conference on Empirical Methods in Natural Language Processing: Industry Track",
    month = nov,
    year = "2025",
    address = "Suzhou (China)",
    publisher = "Association for Computational Linguistics",
    url = "https://aclanthology.org/2025.emnlp-industry.88/",
    doi = "10.18653/v1/2025.emnlp-industry.88",
    pages = "1265--1274",
    ISBN = "979-8-89176-333-3"
}

@misc{json,
    title  = {Introducing {Structured Outputs} in the {API}},
    author = {{OpenAI}},
    year   = {2024},
    url    = {https://openai.com/index/introducing-structured-outputs-in-the-api/},
    note   = {Accessed 2026-06-16}
}

@misc{tool,
    title  = {Tool Calling with {LangChain}},
    author = {{The LangChain Team}},
    year   = {2024},
    url    = {https://www.langchain.com/blog/tool-calling-with-langchain},
    note   = {Accessed 2026-06-16}
}

@misc{cobbe2021trainingverifierssolvemath,
      title={Training Verifiers to Solve Math Word Problems}, 
      author={Karl Cobbe and Vineet Kosaraju and Mohammad Bavarian and Mark Chen and Heewoo Jun and Lukasz Kaiser and Matthias Plappert and Jerry Tworek and Jacob Hilton and Reiichiro Nakano and Christopher Hesse and John Schulman},
      year={2021},
      eprint={2110.14168},
      archivePrefix={arXiv},
      primaryClass={cs.LG},
      url={https://arxiv.org/abs/2110.14168}, 
}

@inproceedings{madaan2023selfrefine,
 author = {Madaan, Aman and Tandon, Niket and Gupta, Prakhar and Hallinan, Skyler and Gao, Luyu and Wiegreffe, Sarah and Alon, Uri and Dziri, Nouha and Prabhumoye, Shrimai and Yang, Yiming and Gupta, Shashank and Majumder, Bodhisattwa Prasad and Hermann, Katherine and Welleck, Sean and Yazdanbakhsh, Amir and Clark, Peter},
 booktitle = {Advances in Neural Information Processing Systems},
 doi = {10.52202/075280-2019},
 editor = {A. Oh and T. Naumann and A. Globerson and K. Saenko and M. Hardt and S. Levine},
 pages = {46534--46594},
 publisher = {Curran Associates, Inc.},
 title = {{Self-Refine}: Iterative Refinement with Self-Feedback},
 url = {https://proceedings.neurips.cc/paper_files/paper/2023/file/91edff07232fb1b55a505a9e9f6c0ff3-Paper-Conference.pdf},
 volume = {36},
 year = {2023}
}

@inproceedings{shinn2023reflexion,
 author = {Shinn, Noah and Cassano, Federico and Gopinath, Ashwin and Narasimhan, Karthik and Yao, Shunyu},
 booktitle = {Advances in Neural Information Processing Systems},
 doi = {10.52202/075280-0377},
 editor = {A. Oh and T. Naumann and A. Globerson and K. Saenko and M. Hardt and S. Levine},
 pages = {8634--8652},
 publisher = {Curran Associates, Inc.},
 title = {Reflexion: language agents with verbal reinforcement learning},
 url = {https://proceedings.neurips.cc/paper_files/paper/2023/file/1b44b878bb782e6954cd888628510e90-Paper-Conference.pdf},
 volume = {36},
 year = {2023}
}

@misc{litellm,
  title        = {{LiteLLM}},
  author       = {{BerriAI}},
  year         = {2023},
  howpublished = {\url{https://github.com/BerriAI/litellm}},
  note         = {Accessed: 2026-06-16}
}

\clearpage
\appendix
\twocolumn[%
\begin{minipage}{\textwidth}
\centering\small
\setlength{\tabcolsep}{4pt}
\setlength{\aboverulesep}{0pt}\setlength{\belowrulesep}{0pt}\renewcommand{\arraystretch}{1.25}
\begin{tabular}{l >{\columncolor{black!7}}c cccc cccc ccc}
\toprule
& & \multicolumn{4}{c}{\textsc{Reflect}} & \multicolumn{4}{c}{\textsc{med}} & \multicolumn{3}{c}{\textsc{Select}} \\
\cmidrule(lr){3-6} \cmidrule(lr){7-10} \cmidrule(lr){11-13}
\noalign{\vskip-\cmidrulewidth}
Model & \multirow{-2}{*}{\textsc{j+h}} & $k{=}2$ & $k{=}3$ & $k{=}4$ & $k{=}5$ & $k{=}2$ & $k{=}3$ & $k{=}4$ & $k{=}5$ & $k{=}3$ & $k{=}4$ & $k{=}5$ \\
\midrule
gpt-oss-20b & 15.9\,s & 2.71$\times$ & 3.37$\times$ & 3.98$\times$ & 4.56$\times$ & 1.21$\times$ & \textbf{1.36$\times$} & 1.44$\times$ & 1.49$\times$ & 2.95$\times$ & \textbf{2.87$\times$} & 3.08$\times$ \\
gpt-oss-120b & 19.3\,s & 2.09$\times$ & 2.76$\times$ & 3.41$\times$ & 4.04$\times$ & 1.08$\times$ & \textbf{1.12$\times$} & 1.15$\times$ & 1.17$\times$ & 1.95$\times$ & \textbf{2.13$\times$} & 2.20$\times$ \\
gpt-5.4-mini & 7.3\,s & 2.04$\times$ & 2.54$\times$ & 3.03$\times$ & 3.52$\times$ & 1.20$\times$ & \textbf{1.27$\times$} & 1.33$\times$ & 1.39$\times$ & 2.08$\times$ & \textbf{2.23$\times$} & 2.41$\times$ \\
\midrule
qwen3.5-9b & 23.3\,s & 2.73$\times$ & 3.75$\times$ & 4.75$\times$ & 5.70$\times$ & 1.14$\times$ & \textbf{1.20$\times$} & 1.29$\times$ & 1.32$\times$ & 3.65$\times$ & \textbf{4.11$\times$} & 4.41$\times$ \\
qwen3.5-27b & 58.5\,s & 2.64$\times$ & 3.45$\times$ & 4.24$\times$ & 5.04$\times$ & 1.13$\times$ & \textbf{1.24$\times$} & 1.28$\times$ & 1.35$\times$ & 4.81$\times$ & \textbf{5.20$\times$} & 5.56$\times$ \\
qwen3.5-122b & 41.5\,s & 3.28$\times$ & 4.26$\times$ & 5.14$\times$ & 6.02$\times$ & 1.19$\times$ & \textbf{1.29$\times$} & 1.41$\times$ & 1.47$\times$ & 6.24$\times$ & \textbf{5.62$\times$} & 5.51$\times$ \\
\bottomrule
\end{tabular}
\captionof{table}{
Average latency per sample of \textsc{j+h} and relative latency multipliers of \textsc{Reflect}, \textsc{med}, and \textsc{Select} with respect to \textsc{j+h}. $k$ denotes the total number of LLM calls. Bold marks the settings recommended in Section~\ref{sec:cost_analysis}.
gpt-5.4-mini is called from OpenAI API. qwen3.5-122b uses four A100 GPUs (80GB), while other local models use a single A100 GPU (80GB).
}
\label{tab:latency}
\end{minipage}
\vspace{1.5em}
]

\section{Latency Analysis}
\label{app:latency}
In addition to the cost analysis in Section~\ref{sec:cost_analysis}, we also measure the latency of the revision methods. Table~\ref{tab:latency} shows the average latency per sample of \textsc{j+h} and relative latency multipliers of \textsc{Reflect}, \textsc{med}, and \textsc{Select} with respect to \textsc{j+h}.
The latency of \textsc{Reflect} increases roughly in proportion to the number of LLM calls. Because \textsc{med} executes its samples in parallel, its latency is determined by the slowest sample. Consequently, its latency is only slightly higher than that of \textsc{j+h} but remains below 1.5$\times$ even as the number of samples increases. \textsc{Select} remains within around 2--3$\times$ for GPT models, whereas for Qwen models, the increased amount of reasoning raises the latency to 3.3–6$\times$. In terms of performance gain relative to the increase in latency $(\frac{\mathrm{gain}}{\mathrm{latency} / \mathrm{latency}_{\textsc{j+h}} - 1})$, \textsc{med} with $k=3$ is the most latency-efficient method on average across all models, quantitatively supporting the guidelines in Section~\ref{sec:cost_analysis}.
Note that the absolute latency of \textsc{j+h} depends heavily on the deployment environment. Consequently, the extent to which the latency multipliers of different test-time compute methods affect practical responsiveness also depends on the deployment environment.
\section{API Pricing}
\label{app:pricing}

Table~\ref{tab:pricing} lists the per-token prices used to compute the API cost in Section~\ref{sec:cost_analysis}. For the local models, we use the prices listed on OpenRouter\footnote{\url{https://openrouter.ai/models}}, and for gpt-5.4-mini, we use the official OpenAI pricing\footnote{\url{https://developers.openai.com/api/docs/models/gpt-5.4-mini}}, both as of June 12, 2026. The cost is simply calculated as the input price times the number of input tokens plus the output price times the number of output tokens.

\begin{center}
\small
\begin{tabular}{l cc}
\toprule
&  \multicolumn{2}{c}{USD per 1M tokens} \\
\cmidrule(lr){2-3}
Model & Input & Output \\
\midrule
gpt-oss-20b   & 0.029 & 0.140 \\
gpt-oss-120b  & 0.039 & 0.180 \\
gpt-5.4-mini  & 0.750 & 4.500 \\
qwen3.5-9b    & 0.100 & 0.150 \\
qwen3.5-27b   & 0.195 & 1.560 \\
qwen3.5-122b  & 0.260 & 2.080 \\
\bottomrule
\end{tabular}
\captionof{table}{Per-token prices used for the cost analysis (as of June 12, 2026).}
\label{tab:pricing}
\end{center}

\begin{table*}[p]
\centering
\small
\begin{tabular}{@{}l p{3.6cm} p{9.4cm}@{}}
\toprule
\textbf{Pattern} & \textbf{Description} & \textbf{Representative scenario (edit $\rightarrow$ cascade)} \\
\midrule
\texttt{arithmetic} & Numeric value feeds a formula chain & \emph{Invoice total chain} (invoice): edit one line's quantity or unit price $\rightarrow$ its line amount, the subtotal, and every charge built on the subtotal recompute in order \\
\addlinespace
\texttt{substitution} & An entity is swapped and its governed attributes are copied in & \emph{Changing an employee role} (org access): the new role's permission groups, approval limit, required trainings, review interval, and support queue replace the old ones \\
\addlinespace
\texttt{add\_remove} & A list element is added/removed and downstream links and aggregates update & \emph{Removing a task} (project schedule): successors relink to the removed task's predecessor, downstream dates pull earlier, and the task count and project end shift \\
\addlinespace
\texttt{threshold} & Effect is conditional on crossing or clamping to a stated threshold & \emph{Crossing a data-volume threshold} (data pipeline): raising the daily volume recomputes the storage estimate and, only if it crosses the threshold, flips the partition strategy and steps the compaction/retention tiers \\
\addlinespace
\texttt{temporal} & Date/time values shift along the dependency chain & \emph{Shifting the trip's anchor date} (travel): every date derived from the anchor by a fixed offset shifts by the same delta, while independent dates stay \\
\addlinespace
\texttt{status\_flip} & A state change moves a set of status-governed fields together & \emph{Marking the invoice as paid} (invoice): every status-governed field moves to its paid-state value at once \\
\bottomrule
\end{tabular}
\caption{The six propagation patterns covered by RevPropBench's 50 scenarios, with a representative scenario for each. Every pattern appears across multiple domains.}
\label{tab:scenarios}
\end{table*}

\begin{figure*}[p]
\centering
\begin{subfigure}[t]{\textwidth}
\small
\begin{tcolorbox}[colback=white,colframe=black!55,boxrule=0.5pt,arc=2pt,left=4pt,right=4pt,top=3pt,bottom=3pt]
\noindent\textbf{Conversation pattern.} Center the artifact on a trip anchored on one date, where several other dates each derive from that anchor by a fixed offset stated in the conversation, so the anchor alone determines them. Around this core, add the realistic, independent details a traveler would record that do not depend on the dates, and vary their number from sample to sample.

\smallskip
\noindent\textbf{Revision pattern.} Move the anchor start date to a specific new date a few days from the original.

\smallskip
\noindent\textbf{Propagation pattern.} Every date that derives from the anchor shifts by the same delta, since each is the anchor plus a fixed offset, so the new anchor uniquely fixes them all. Values that do not derive from the anchor date stay unchanged.
\end{tcolorbox}
\caption{Scenario (domain: \texttt{travel\_itinerary}, theme: ``Shifting the trip's anchor date'', pattern: \texttt{temporal}).}
\label{fig:example-skel}
\end{subfigure}

\par\bigskip

\begin{subfigure}[t]{\textwidth}
\begin{tcolorbox}[colback=white,colframe=black!55,boxrule=0.5pt,arc=2pt,left=4pt,right=4pt,top=3pt,bottom=3pt]
{\small\noindent\textbf{Conversation:}}
\begin{lstlisting}[language=json, basicstyle=\fontsize{7}{9}\selectfont\ttfamily, aboveskip=2pt, belowskip=2pt, breakindent=0pt, breakautoindent=false]
[user]      I'm planning a trip that begins on May 14, 2026.
[assistant] [{"op": "add", "path": "/trip", "value": {"start_date": "2026-05-14"}}]
[user]      Plan for me to fly out two days after the trip begins.
[assistant] [{"op": "add", "path": "/trip/outbound_flight_date", "value": "2026-05-16"}]
[user]      Have me check into the hotel three days after the trip begins.
[assistant] [{"op": "add", "path": "/trip/hotel_check_in_date", "value": "2026-05-17"}]
[user]      Set my flight home for eight days after the trip begins.
[assistant] [{"op": "add", "path": "/trip/return_flight_date", "value": "2026-05-22"}]
[user]      Put the guided food tour five days after the trip begins.
[assistant] [{"op": "add", "path": "/trip/guided_food_tour_date", "value": "2026-05-19"}]
[user]      Schedule the museum visit for six days after the trip begins.
[assistant] [{"op": "add", "path": "/trip/museum_visit_date", "value": "2026-05-20"}]
[user]      Make the hotel checkout seven days after the trip begins.
[assistant] [{"op": "add", "path": "/trip/hotel_check_out_date", "value": "2026-05-21"}]
[user]      Add a nice dinner reservation four days after the trip begins.
[assistant] [{"op": "add", "path": "/trip/dinner_reservation_date", "value": "2026-05-18"}]
[user]      Add a reminder to call the hotel one day before I check in.
[assistant] [{"op": "add", "path": "/trip/hotel_call_reminder", 
              "value": {"date": "2026-05-16", "note": "Call the hotel to confirm the reservation"}}]
\end{lstlisting}
\smallskip
{\small\noindent\textbf{Revision request:} \emph{``Please change the trip to begin on May 17, 2026.''}}
\end{tcolorbox}
\caption{Generated sample.}
\label{fig:example-sample}
\end{subfigure}
\caption{An example scenario and its corresponding generated sample.}
\label{fig:example}
\end{figure*}

\begin{figure*}[t]
\centering
\includegraphics[width=\linewidth]{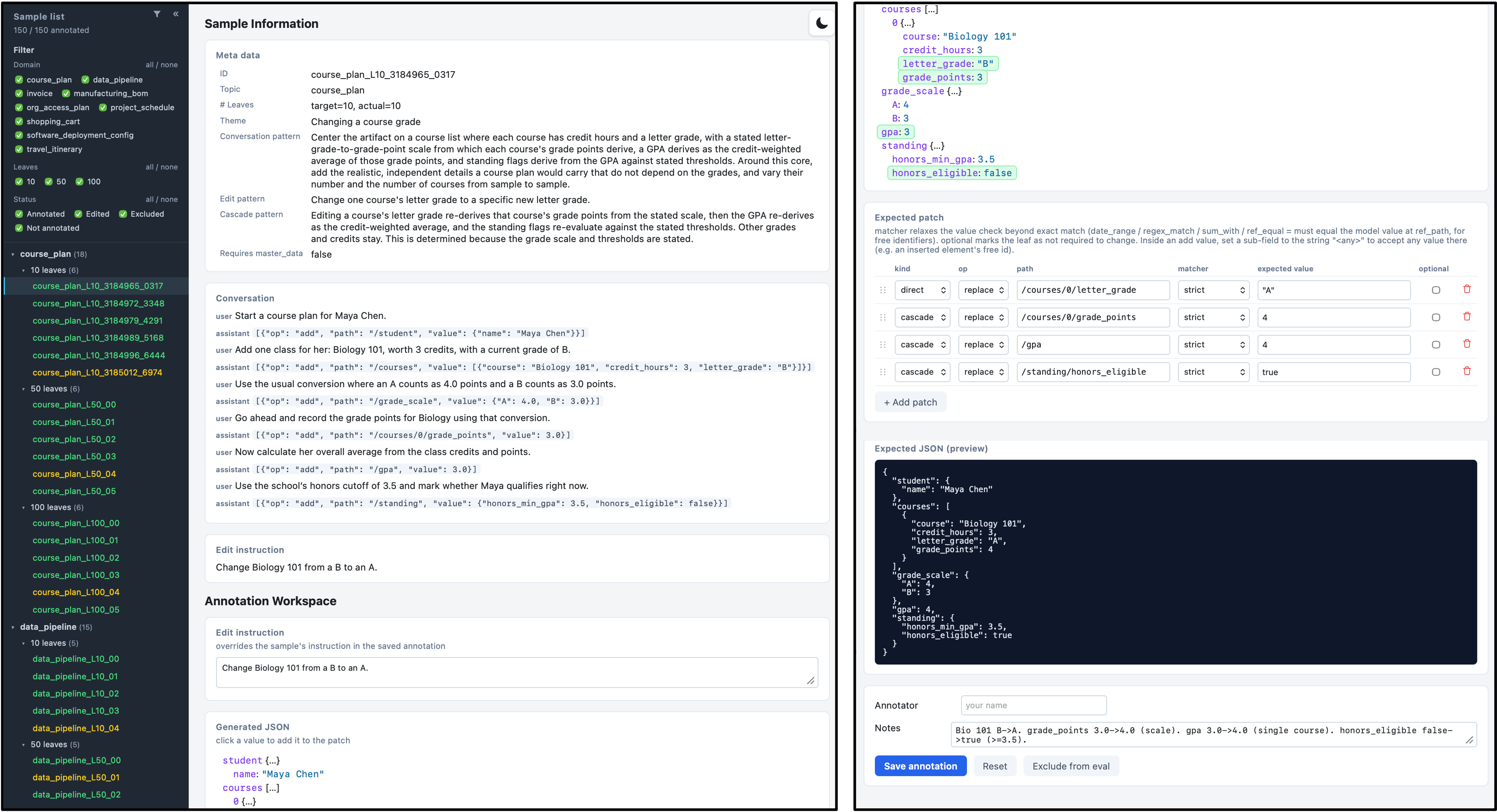}
\caption{Graphical user interface of the annotation tool.}
\label{fig:gui}
\end{figure*}

\section{Details of Benchmark Construction}
\subsection{Scenarios}
The 50 scenarios cover nine domains and six propagation patterns.
Table~\ref{tab:scenarios} summarizes the six patterns with a representative scenario for each.
Figure~\ref{fig:example} shows a scenario and a sample generated from it. The sample includes a conversation and a revision request.
This example simulates a user planning a trip schedule.
The user specifies activities relative to the trip date, and the LLM records them.
The user then requests a shift in the trip start date.
To satisfy this revision request, the LLM must update the dates of the relevant entries accordingly. However, in this case, the required updates are not uniquely determined by the JSON artifact alone.
A simple interpretation is to shift the entire schedule. In practice, however, the user may want to keep certain dates fixed, which cannot be inferred from the artifact alone.
The dependency is identifiable only when the user describes activities in relation to the trip start date, such as “three days after the trip begins.” Only then is it clear that changes to the trip start date should be propagated to the related activities.
In cases like this, when an artifact is created through a conversation, the conversation may contain information that defines dependencies. This sample successfully simulates that situation.

\subsection{Annotation}
\label{app:annotation}
\paragraph{GUI tool}
Figure~\ref{fig:gui} shows the GUI of our annotation tool.
Annotators can select a sample to annotate from the sidebar. The main panel displays the metadata, conversation history, and revision request for the selected sample. While reviewing this information, annotators perform the annotation in the Annotation Workspace. For the revision request, annotators generally do not modify it. However, if the request is ambiguous and makes it difficult to assign a unique annotation, they are allowed to revise the request to make it more specific.

The main annotation task is the creation of the gold patch, which is performed in the Expected Patch section.
Here, annotators add patch operations consisting of \texttt{op} (\texttt{replace}, \texttt{add}, or \texttt{remove}), \texttt{path}, and \texttt{expected value}.
For each operation, annotators also specify its \texttt{kind}: \texttt{direct} if the target is explicitly mentioned in the revision request, or \texttt{cascade} if it is not explicitly mentioned but must be revised due to a dependency.
For value comparison, annotators can choose among several matching methods, including exact matching (\texttt{strict}), regular-expression matching (\texttt{regex\_match}), accepting any value (\texttt{any}), and checking equality with the value of another leaf element (\texttt{ref\_equal}). 
If an operation is contextually valid both when applied and when left unchanged, annotators can mark it as optional by checking the \texttt{optional} box.

The Expected JSON (preview) displays the JSON artifact after the patch has been applied, allowing annotators to visually inspect the resulting artifact.
Finally, at the bottom of the main panel, annotators enter their name and any notes, then click Save to complete the annotation for one sample.

\paragraph{Process}
Claude Code (Opus 4.8) generates tentative gold patches in batches by artifact size. First, Claude Code generates all patches for artifact size 10, and the annotator then reviews and revises the entire batch. The same process is then repeated for artifact size 50, followed by artifact size 100.
For each subsequent batch, Claude Code refers to the annotator’s revisions from the previous batch when generating new tentative gold patches.
During the human review, when the annotator identifies points that required further clarification, they ask Claude Code about the rationale or background behind the tentative patch. If the explanation is judged to be reasonable, the patch is kept. If Claude Code has misunderstood any aspect of the annotation, the annotator corrects the patch accordingly.
Here, the human annotator uses the GUI, while Claude Code annotates patches via CLI.
\paragraph{Results}
One annotator carefully performed the annotation and subsequent double-checking with assistance from Claude Code (Opus 4.8), spending more than 32 hours over 10 days.
23 out of 150 samples (15\%) were revised by the annotator: one involved a revision to the revision request, and 22 involved revisions to the patches.
The actual revisions can be checked in the GUI of the provided annotation tool.
\section{Details of Merging Rules}
\label{app:methods}

For \textsc{or}, \textsc{and}, \textsc{maj}, and \textsc{med}, each of the $k$ sampled patches is first applied to the current artifact, yielding $k$ candidate artifacts.
We compare these candidate artifacts rather than the patches themselves, as the same edit can be expressed by many different JSON patches.

\textsc{or}, \textsc{and}, and \textsc{maj} merge the candidates element-wise.
Each leaf-level key--value pair in a JSON object is treated as an independent merge unit.
If the value is an array, elements at the same index are treated as separate merge units only when the current artifact and all candidates have the same length.
Otherwise, the key--array pair is treated as a single, indivisible merge unit.
For each merge unit, we compare whether each candidate adds, modifies, removes, or leaves it unchanged, based on the presence of the key and its corresponding value.
\textsc{and} adopts a change only when all candidates propose the same change.
\textsc{maj} adopts a change only when a strict majority ($>k/2$) propose the same change.
\textsc{or} adopts a change whenever at least one candidate proposes one (if multiple different changes are proposed, the change from the earliest sampled candidate is adopted).
The merged artifact is then converted back into a JSON patch.
If the merged artifact exactly matches one of the candidate artifacts, we reuse that candidate's original patch.
Otherwise, a new patch is constructed from the difference between the current and merged artifacts.

In contrast, \textsc{med} returns one whole candidate. 
Each candidate artifact is flattened into its leaf-level merge
units, and the disagreement between two candidates is measured as the fraction of mismatched merge units among all units appearing in either candidate (a unit missing
from one of them also counts as a mismatch). 
For each candidate, this disagreement is averaged over the other $k-1$ candidates, and the candidate with the minimum mean disagreement is selected.
\section{Safeguards for Critical Revisions}

In practical use cases such as financial data processing and invoice processing, even a single over-edit or miss in revision propagation can result in significant errors.
Since our problem setting assumes interactive conversation between humans and LLMs, we suggest incorporating human review as a safeguard in such cases.
For example, a practical safeguard would be for the LLM to present all propagated changes before application and apply only those explicitly approved by the user. We believe that presenting the required edits alone can substantially reduce human effort. Additionally, we could consider several approaches to reduce the effort required for human review, such as using an LLM-as-a-judge to prioritize edits based on their criticality or allowing users to specify critical items in advance. Note that these suggestions are our conceptual ideas and have not yet been experimentally validated.
\section{System Prompts}
\label{app:prompts}
In this section, we provide the prompts used for both data generation and evaluation.
Figures~\ref{prompt:gen-system}--\ref{prompt:gen-edit} show the prompts for data generation.
Curly-brace tokens such as \textcolor{blue}{\texttt{\{domain\_name\}}} are placeholders filled with specific values.
The system prompt (Figure~\ref{prompt:gen-system}) and the first-turn user prompt (Figure~\ref{prompt:gen-turn1}) are first provided as the initial context, from which the first user--LLM conversation turn is generated. 
Subsequently, at each turn, the subsequent-turn user prompt (Figure~\ref{prompt:gen-turnk}) is appended to the current context to generate the next conversation turn. Both the prompt and the generated turn are then retained in the context.
This process is repeated until the full conversation has been generated. Finally, the revision-instruction prompt (Figure~\ref{prompt:gen-edit}) is appended to the end of the context to generate a revision request. 

Figures~\ref{prompt:rev-system}--\ref{prompt:rev-reflect} show prompts for evaluation.
For a single inference, the revision system prompt (Figure~\ref{prompt:rev-system}) and the revision user prompt (Figure~\ref{prompt:rev-user}) are provided as the context to generate a JSON patch. In Figure~\ref{prompt:rev-user}, \textsc{j} includes only \textcolor{blue}{\texttt{\{current\_json\}}}, \textsc{h} includes only \textcolor{blue}{\texttt{\{conversation\}}}, and \textsc{j+h} includes both, while \textcolor{blue}{\texttt{\{requested\_edit\}}} is always included.
For parallel sampling, multiple JSON patches are generated in parallel using the same system prompt and user prompt.
For \textsc{Select}, multiple JSON patches are first generated, after which the final patch is selected using the system prompt and user prompt for \textsc{Selection} (Figures~\ref{prompt:rev-select} and~\ref{prompt:rev-select-user}).
For \textsc{Reflect}, after a single inference, the user prompt for \textsc{Reflect} (Figure~\ref{prompt:rev-reflect}) is appended to the context to generate a new JSON patch. This process is repeated for $k-1$ iterations.

\lstdefinestyle{promptstyle}{
  basicstyle=\ttfamily\footnotesize,
  breaklines=true,
  breakautoindent=false,
  breakindent=0pt,
  columns=fullflexible,
  keepspaces=true,
  showstringspaces=false,
  aboveskip=0pt,
  belowskip=0pt,
  moredelim=*[is][\color{blue!80!black}]{`}{`},
}
\tcbset{
  promptbox/.style={
    enhanced,
    breakable=false,
    colback=white,
    colframe=black!55,
    colbacktitle=black,
    coltitle=white,
    boxrule=0.5pt,
    titlerule=0pt,
    arc=2pt,
    left=4pt, right=4pt, top=3pt, bottom=3pt,
    toptitle=2pt, bottomtitle=2pt,
    fonttitle=\ttfamily\bfseries\footnotesize,
  },
  promptboxgen/.style={
    promptbox,
    colback=cyan!2,
    colframe=cyan!60!black,
    colbacktitle=cyan!32,
    coltitle=black,
  },
  promptboxeval/.style={
    promptbox,
    colback=green!2,
    colframe=green!45!black,
    colbacktitle=green!28,
    coltitle=black,
  },
}
\renewcommand{\dbltopfraction}{0.95}
\renewcommand{\dblfloatpagefraction}{0.85}

\begin{figure*}[tp]
\begin{tcblisting}{promptboxgen, title={Data generation: system prompt}, listing only, listing options={style=promptstyle}}
You are generating data samples in a benchmark for evaluating the ability of LLMs to propagate feedback across dependent elements in JSON objects previously generated by the LLMs themselves.
The benchmark consists of two phases: generation phase and revision phase. In the generation phase, muliple JSON objects are iteratively generated through multi-turn conversations between a 
user and an LLM. In the subsequent revision phase, we evaluate whether the LLM can propagate user-specified modifications across all dependent elements of the generated JSON objects.

You first generate a multi-turn synthetic conversation step by step, one turn at a time, as pairs
consisting of the user's instruction and the LLM's response (JSON format).
After generating the full conversation between the user and the LLM, you then generate a revision 
instruction that the user gives to the LLM. The revision instruction should specify ONLY a single 
leaf element of the JSON object to measure whether the LLM can autonomously identify and revise the other necessary parts.
The pair of the multi-turn conversation and the revision instruction becomes one sample.

# Sample generation guideline
Here, you will generate one sample that satisfies the requirements specified below.
## Domain
`{domain_name}`

## Theme
`{theme}`

## Conversation pattern
`{conversation_pattern}`

## Edit instruction pattern
`{edit_pattern}`

## Cascade pattern
`{cascade_pattern}`

# Rules
- The conversation between the user and the LLM must read as natural: ordinary user requests 
  building up a realistic artifact.
- Whatever the revision instruction should cause to change must be uniquely determined: for the 
  edited field and every value that must update in response, there is exactly ONE correct new 
  value, with no room for more than one valid interpretation.
- A value that aggregates or summarizes other parts of the artifact may be generated only after 
  every value it depends on is already present. After such a value has been generated, later 
  turns must add only values that do not affect it -- never add or change anything it depends on.
\end{tcblisting}
\caption{System prompt for data generation.}
\label{prompt:gen-system}
\end{figure*}

\begin{figure*}[tp]
\begin{tcblisting}{promptboxgen, title={Data generation: first-turn user prompt}, listing only, listing options={style=promptstyle}}
Let's generate a sample for the specified guideline. The complete conversations should hold `{target_leaves}` primitive leaf values of JSON objects in total, built up over the turns.
For the first turn, output the pair of the user's generation instruction and the corresponding JSON output from the LLM in the following format.
{
  "user_instruction": "<a natural opening request, phrased the way a real user would talk to an 
                       assistant. Introduce just ONE foundational piece of information.'>",
  "json_output": [{"op": "add", "path": "<JSON Pointer>", "value": <the value to add>}]
}
The user_instruction must read like a genuine user message -- never mention "key", "field", "JSON", or value types, and never name the JSON key literally. The json_output is a JSON Patch that ADDS this turn's new content to the artifact being built: a list of operations. Introduce a new field or collection by adding it at its own pointer; when the value is a collection, add it whole with its first element. The json_output may add at most `{max_leaves_per_turn}` primitive leaf values to the artifact this turn (counted across any nested objects and array elements).
\end{tcblisting}
\caption{User prompt for the first conversation turn.}
\label{prompt:gen-turn1}
\end{figure*}

\begin{figure*}[tp]
\begin{tcblisting}{promptboxgen, title={Data generation: subsequent-turn user prompt}, listing only, listing options={style=promptstyle}}
So far the conversation holds `{leaves_so_far}` of `{target_leaves}` target primitive leaf values (i.e., remaining leaf budget: `{remaining_leaves}`); keep track of the remaining number of leaves and adjust the conversation so that it concludes within that limit.
For the next turn, output the pair of the user's generation instruction and the corresponding JSON output from the LLM in the following format.
{
  "user_instruction": "<a natural follow-up request, phrased the way a real user actually talks,
                       that adds ONE new piece of information related to what already exists so 
                       that changing it later would cascade.>",
  "json_output": [{"op": "add", "path": "<JSON Pointer>", "value": <value>}]
}
The user_instruction must read like a genuine user message -- never mention "key", "field", "JSON", or value types, and never name the JSON key literally. The json_output is a JSON Patch that ADDS this turn's new content: a list of operations. Introduce a new field or collection by adding 
it at its own pointer; append another element to a collection that already exists with a pointer ending in "/-". Do NOT re-add or restate anything already present in the artifact -- add only this turn's new content. Every added value must be logically consistent with all values already in the conversation. The json_output may add at most `{max_leaves_per_turn}` NEW primitive leaf values to the artifact this turn (counted across any nested objects and array elements). Only on the final turn, if the remaining leaf budget and this per-turn leaf limit are too small to complete the conversation, you may exceed them by one or two leaves.
\end{tcblisting}
\caption{User prompt for every subsequent conversation turn.}
\label{prompt:gen-turnk}
\end{figure*}

\begin{figure*}[tp]
\begin{tcblisting}{promptboxgen, title={Data generation: revision-instruction prompt}, listing only, listing options={style=promptstyle}}
The generation phase is complete. The full artifact is:
`{generated_jsons}`

Now produce the revision instruction the user gives next, carrying out the edit described in the guideline:
`{edit_pattern}`

The instruction must:
- Target EXACTLY ONE element -- a single primitive leaf (number, string, date, boolean), or the 
  addition or removal of EXACTLY ONE array element; never a whole object or array.
- Give a concrete new value (a specific scalar), not a directional hint.
- Be chosen so that every dependent field has ONE uniquely-correct new value.
- Read like a genuine user message -- never mention "key", "field", "JSON", or value types.

The output format is as follows:
{
    "edit_instruction": "<the single-target revision instruction in natural language, phrased as 
                         a real user request>"
}
\end{tcblisting}
\caption{User prompt for generating a revision request.}
\label{prompt:gen-edit}
\end{figure*}

\begin{figure*}[tp]
\begin{tcblisting}{promptboxeval, title={Evaluation: system prompt}, listing only, listing options={style=promptstyle}}
You are a revision agent for a JSON artifact that is edited over multiple turns.

You are given context for the artifact -- the conversation that built it and/or its current state -- and a single requested edit (and, when the edit introduces a new entity, reference data for that entity). Compute the full consequence of the edit and apply it: change the field the edit changes directly, and every other field whose value is FORCED to change because it depends on a field that the edit changes -- directly or through a chain of such dependencies.

A field is FORCED to change in any of these ways, and you must follow all of them to the end of the chain:
  - it is computed or derived from a changed field (do the arithmetic; pick the correct branch 
    of any conditional such as a threshold crossing or a clamp to a bound, and write the final 
    value);
  - it NAMES, REFERS TO, ROUTES TO, or is ASSIGNED TO an entity that the edit changes (an owner, 
    source, assignee, target, identifier, or label) -- update the referent to the new entity even 
    when no number changes;
  - it is a control, policy, status, or routing field that is GATED by a value which, after the edit, crosses a threshold or moves into a different band, tier, or category -- switch it to the branch the new value selects.

Aim for a revision that is both COMPLETE and EXACT:
  - Complete: follow every dependency chain -- value, reference, and gating -- to its end; do not 
    stop at the first level, and do not stop at numeric fields.
  - Exact: change a field only if its value genuinely becomes DIFFERENT. For every candidate, 
    read its current value and compare it to the value you computed; if they are equal, leave it
    unchanged. Do not change a field merely because it is topically related.

Before returning, review the operations once: drop any operation whose value equals the current value, and add any forced change -- value, reference, or gated -- that you missed.

# Output format
Produce a JSON Patch (RFC 6902): a JSON array of operations that transforms the
current artifact into the corrected version. The operations are applied strictly
in order, each to the result of the previous one: when an array element is added
or removed, the indices of the later elements shift accordingly, so every path
must address the artifact state at that point in the sequence. Apply the
requested edit AND change every other field that must change to keep the artifact
internally consistent; do NOT modify unrelated fields. Each operation has the form
{"op": "replace" | "add" | "remove", "path": "<RFC 6901 JSON Pointer>", "value": <new value>}
("value" is omitted for "remove"). Output ONLY the JSON array -- no prose, no code fences.
\end{tcblisting}
\caption{Revision system prompt, shared by all revision methods.}
\label{prompt:rev-system}
\end{figure*}

\begin{figure*}[tp]
\begin{tcblisting}{promptboxeval, title={Evaluation: user prompt}, listing only, listing options={style=promptstyle}}
Conversation that built the artifact (shows how each value was derived):
`{conversation}`

Current JSON artifact:
`{current_json}`

Requested edit:
`{requested_edit}`
\end{tcblisting}
\caption{Revision user prompt; the conversation (\textsc{h}) and JSON (\textsc{j}) blocks are included per baseline.}
\label{prompt:rev-user}
\end{figure*}

\begin{figure*}[tp]
\begin{tcblisting}{promptboxeval, title={Evaluation: system prompt for \textsc{Select}}, listing only, listing options={style=promptstyle}}
You are the selector for a revision task on a JSON artifact that is edited over multiple turns.

You are given the conversation that built the artifact, the current JSON artifact, a single requested edit (and, when the edit introduces a new entity, reference data for that entity), and several candidate JSON Patches produced independently for that edit. Exactly one candidate must be chosen.

Judge each candidate against the artifact and the requested edit: the right candidate applies the 
requested edit and changes every field whose value is FORCED to change as a consequence -- values 
computed or derived from a changed field, fields that name or refer to a changed entity, and control, policy, status, or routing fields gated by a value that crosses a threshold -- while leaving every other field untouched. Verify the arithmetic of changed values, check that no forced change is missing, and check that nothing unrelated is modified.

Output ONLY the number of the chosen candidate. No prose, no punctuation.
\end{tcblisting}
\caption{System prompt for \textsc{Select}.}
\label{prompt:rev-select}
\end{figure*}

\begin{figure*}[tp]
\begin{tcblisting}{promptboxeval, title={Evaluation: user prompt for \textsc{Select}}, listing only, listing options={style=promptstyle}}
`{revision user message: conversation, current JSON, and requested edit}`

Candidate patches:

Candidate 1:
`{patch_1}`

Candidate 2:
`{patch_2}`

...
\end{tcblisting}
\caption{User prompt for \textsc{Select}.}
\label{prompt:rev-select-user}
\end{figure*}

\begin{figure*}[tp]
\begin{tcblisting}{promptboxeval, title={Evaluation: user prompt for \textsc{Reflect}}, listing only, listing options={style=promptstyle}}
Here is the artifact after applying the JSON Patch you just produced:

# Artifact after applying your patch
`{applied}`

Review this artifact against the user feedback. Treat your patch as correct by default; you are looking only for definite errors, of these kinds:
1. a field the edit forces to change (directly or through a dependency chain) that still shows 
   its old value;
2. a field that was modified although nothing forces it to change (revert it);
3. a value that contradicts how the conversation and the artifact determine it -- correct a value 
   only if you can name the specific mistake in it; a re-derivation that merely differs from the 
   current value is not evidence of a mistake, so in that case keep the current value;
4. an operation from your patch whose change is absent above because its path did not address an 
   existing field (re-apply it with a valid path).

Then output a corrective JSON Patch containing ONLY the operations for the definite errors you found, applied IN ORDER to the artifact shown above (the result of your previous patch), so every 
path must address that artifact. If you found no definite error, output an empty array: []. Output ONLY the JSON array -- no prose, no code fences.
\end{tcblisting}
\caption{User prompt for \textsc{Reflect}}
\label{prompt:rev-reflect}
\end{figure*}

\end{document}